%% file: main.tex
\documentclass{article}
\usepackage[utf8]{inputenc}
\usepackage{main}
\usepackage{microtype}
\usepackage{graphicx}
\usepackage{times}
\usepackage{amsmath}
\usepackage{amssymb}
\usepackage{enumitem}
\usepackage{booktabs}
\usepackage{natbib}
\definecolor{mydarkblue}{rgb}{0,0.08,0.45}
\usepackage[colorlinks=true,linkcolor=mydarkblue,citecolor=mydarkblue,filecolor=mydarkblue,urlcolor=mydarkblue]{hyperref}
\usepackage{fancyhdr}

\usepackage{xspace}
\usepackage{float}
\usepackage{multirow}

\usepackage{url}
\usepackage{xcolor}
\usepackage[most]{tcolorbox}
\usepackage{subcaption}
\usepackage{tikz}
\usepackage{pgfplots}
\usepackage{wrapfig}
\usepackage{tabularx}
\usepackage{caption}
\usepackage{amsfonts}
\usepackage{nicefrac}
\usepgfplotslibrary{groupplots}
\usetikzlibrary{matrix}
\pgfplotsset{compat=1.18}

\setlist[itemize,1]{leftmargin=10pt}

\input{commands}

\definecolor{YaleBlue}{RGB}{0, 53, 107}
\definecolor{UNCBlue}{RGB}{75, 156, 211}
\definecolor{PennRed}{RGB}{153, 0, 0}

\newcommand{\Yale}{\hspace{.1em}^{\textcolor{YaleBlue}{\boldsymbol{Y}}}}
\newcommand{\UNC}{\hspace{.1em}^{\textcolor{UNCBlue}{\boldsymbol{C}}}}
\newcommand{\UPenn}{\hspace{.1em}^{\textcolor{PennRed}{\boldsymbol{P}}}}

\newcommand{\github}{\raisebox{-1.5pt}{\includegraphics[height=1.05em]{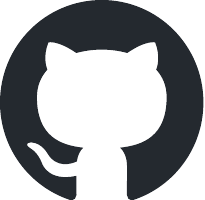}}\xspace}

\usepackage[flushmargin,hang]{footmisc}
\newtcolorbox{taskexample}[1]{
    colback=black!1,
    colframe=black!30,
    boxrule=0.4pt,
    arc=2pt,
    left=5pt,
    right=5pt,
    top=5pt,
    bottom=5pt,
    before skip=0.8em,
    after skip=0.8em,
    title={\textbf{#1}},
    fonttitle=\small,
    coltitle=black,
    colbacktitle=black!6
}

\newcommand{\taskfield}[1]{\vspace{0.25em}\noindent\textbf{#1.}}
\newlist{compactitem}{itemize}{1}
\setlist[compactitem,1]{
    label=\textbullet,
    leftmargin=1.2em,
    itemsep=0.15em,
    topsep=0.15em
}

\definecolor{darkblue}{rgb}{0.0, 0.0, 0.55}

\renewcommand{\eg}{\hbox{\emph{e.g.,}}\xspace}

\newcommand{\nexample}{1,000\xspace}
\newcommand{\napp}{33\xspace}

\newcommand{\ours}{OpenComputer\xspace}

\newcommand{\answerTODO}[1][]{\textcolor{red}{\bf [TODO]}}

\begin{document}

\title{\ours: Verifiable Software Worlds for\\ Computer-Use Agents}

\author{
\textbf{Jinbiao Wei}$\Yale$ \;
\textbf{Qianran Ma}$\UPenn$ \;
\textbf{Yilun Zhao}$\Yale$ \;
\textbf{Xiao Zhou}$\Yale$ \;
\textbf{Kangqi Ni}$\UNC$ \;
\textbf{Guo Gan}$\Yale$ \;
\textbf{Arman Cohan}$\Yale$ \\ [7pt]
$\Yale$Yale NLP Lab \quad $\UPenn$University of Pennsylvania \quad $\UNC$University of North Carolina at Chapel Hill \\ [4pt]
\github\url{https://github.com/echo0715/OpenComputer}  \\ [4pt]
Correspondence to: \href{mailto:jinbiao.wei@yale.edu}{jinbiao.wei@yale.edu}, \href{mailto:yilun.zhao@yale.edu}{yilun.zhao@yale.edu}
}

\maketitle
\thispagestyle{fancy}
\fancyhf{}
\fancyhead[L]{%
    \raisebox{-6pt}{%
        \includegraphics[height=1.1cm]{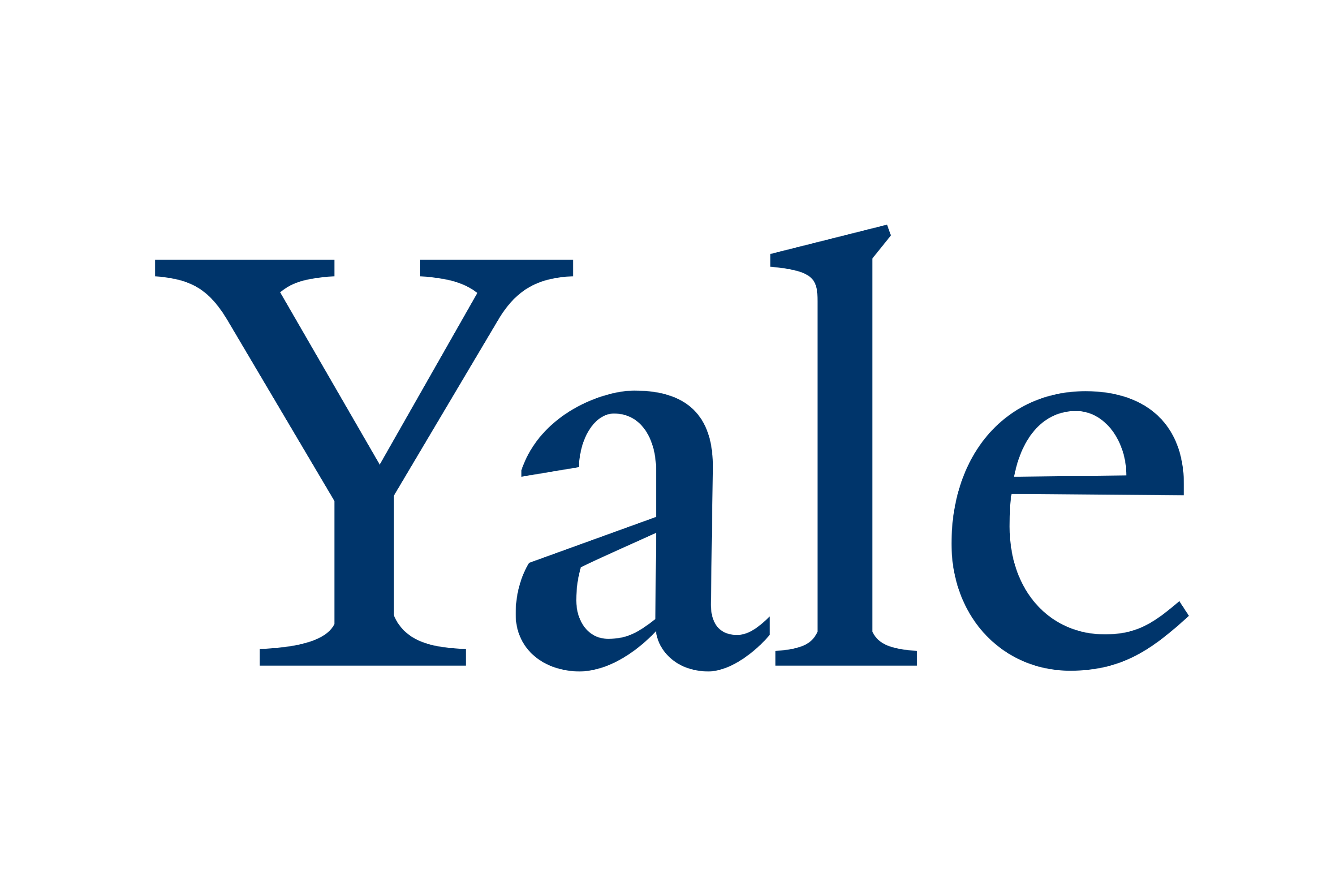}\hspace{0.1cm}%
        \raisebox{0.2cm}{\includegraphics[height=0.65cm]{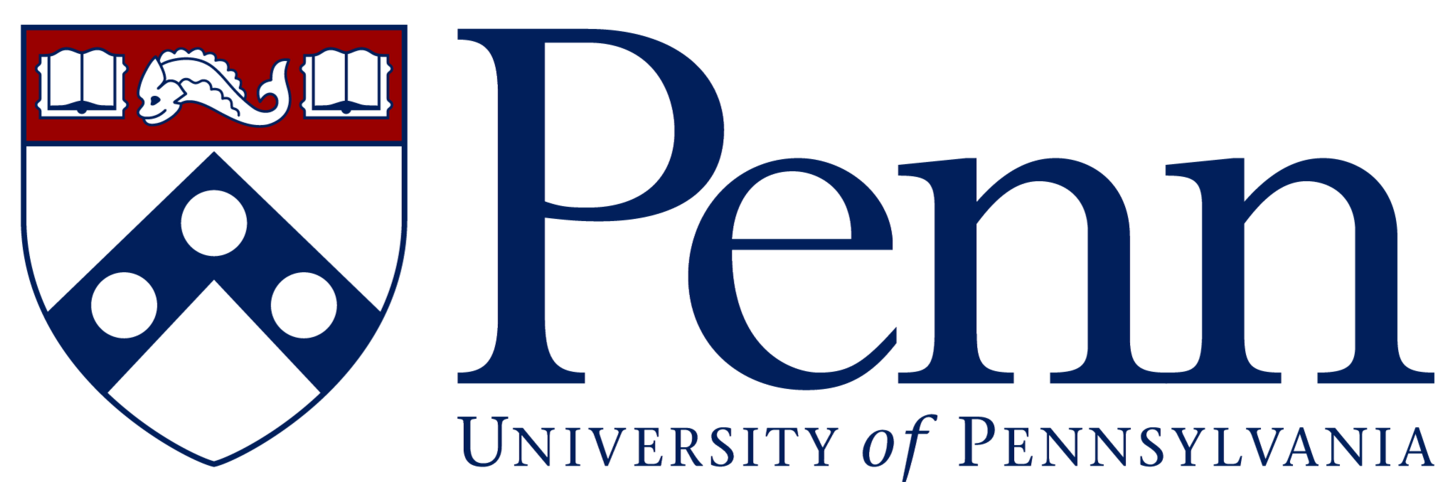}}\hspace{0.35cm}%
        \raisebox{0.1cm}{\includegraphics[height=0.9cm]{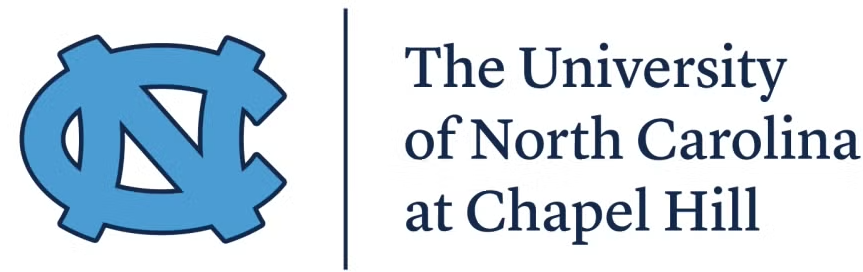}}\hspace{0.3cm}%
    }%
}
\fancyfoot[C]{\thepage}
\renewcommand{\headrulewidth}{0pt}
\setlength{\headheight}{37pt}
\addtolength{\topmargin}{-7pt}
\setlength{\headsep}{0mm}

\vspace{-1.0em}

\pagestyle{plain}

\begin{abstract}
\input{sections/0-abstract}
\end{abstract}

\input{sections/introduction}
\input{sections/related_work}
\input{sections/methodology}
\input{sections/experiment}
\input{sections/analysis}
\input{sections/conclusion}
\input{Appendix/limitations}

\bibliographystyle{plainnat}
\bibliography{references}

\newpage

\appendix
\input{Appendix/repair_module}
\input{Appendix/llm_as_judge_problem}
\input{Appendix/task_example}

\end{document}

%% file: commands.tex
\usepackage{tikz}
\usepackage{soul}
\usepackage{colortbl}
\usepackage{subcaption}
\usepackage{bbm}
\usepackage{adjustbox}
\usepackage{makecell}
\usepackage[framemethod=TikZ]{mdframed}

\definecolor{mypositive}{RGB}{0, 128, 0}
\definecolor{mynegative}{RGB}{220, 20, 60}
\definecolor{mypositive}{RGB}{24, 103, 173}
\definecolor{mynegative}{RGB}{255, 128, 0}

\newcommand{\eg}{\emph{e.g.,}\xspace}

\definecolor{customblue}{HTML}{1F77B4}
\definecolor{customorange}{HTML}{FF7F0E}
\definecolor{customgreen}{HTML}{2CA02C}
\definecolor{no}{HTML}{BA8E23}
\definecolor{darkgreen}{RGB}{0,100,0}
\definecolor{darkred}{RGB}{139,0,0}

\newcommand{\deltaScore}[1]{%
\begingroup
\ifdim #1pt > 0pt
\textcolor{darkred}{\scriptsize$^{#1}$}%
\else
\textcolor{darkgreen}{\scriptsize$^{#1}$}%
\fi
\endgroup
}

\newtcolorbox{boxblue}{enhanced,colback=blue!5!white,colframe=blue!75!black,breakable=true}

\usepackage{listings}
\usepackage{wrapfig}

\lstdefinelanguage{json}{
    basicstyle=\scriptsize\ttfamily,
    numbers=none,
    numberstyle=\tiny,
    stepnumber=1,
    numbersep=5pt,
    showstringspaces=false,
    breaklines=true,
    frame=none,
    backgroundcolor=\color{white},
    literate=
     *{:}{{{\color{black}{:}}}}{1}
      {,}{{{\color{black}{,}}}}{1}
      {\{}{{{\color{black}{\{}}}}{1}
      {\}}{{{\color{black}{\}}}}}{1}
      {[}{{{\color{black}{[}}}}{1}
      {]}{{{\color{black}{]}}}}{1},
}

%% file: sections/0-abstract.tex
We present \textbf{\ours}, a verifier-grounded framework for constructing verifiable software worlds for computer-use agents. \ours integrates four components: 
(1) app-specific state verifiers that expose structured inspection endpoints over real applications, 
(2) a self-evolving verification layer that improves verifier reliability using execution-grounded feedback, 
(3) a task-generation pipeline that synthesizes realistic and machine-checkable desktop tasks, and 
(4) an evaluation harness that records full trajectories and computes auditable partial-credit rewards. In its current form, \ours covers 33 desktop applications and 1,000 finalized tasks spanning browsers, office tools, creative software, development environments, file managers, and communication applications. Experiments show that OpenComputer's hard-coded verifiers align more closely with
human adjudication than LLM-as-judge evaluation, especially when success depends on fine-grained application state. Frontier agents struggle with end-to-end completion despite partial progress, and open-source models exhibit sharp drops from their OSWorld-Verified scores, exposing a persistent gap in robust computer automation.

\vspace{-10pt}

%% file: sections/introduction.tex
\begin{figure*}[h]
    \centering
    \includegraphics[width=0.95\textwidth]{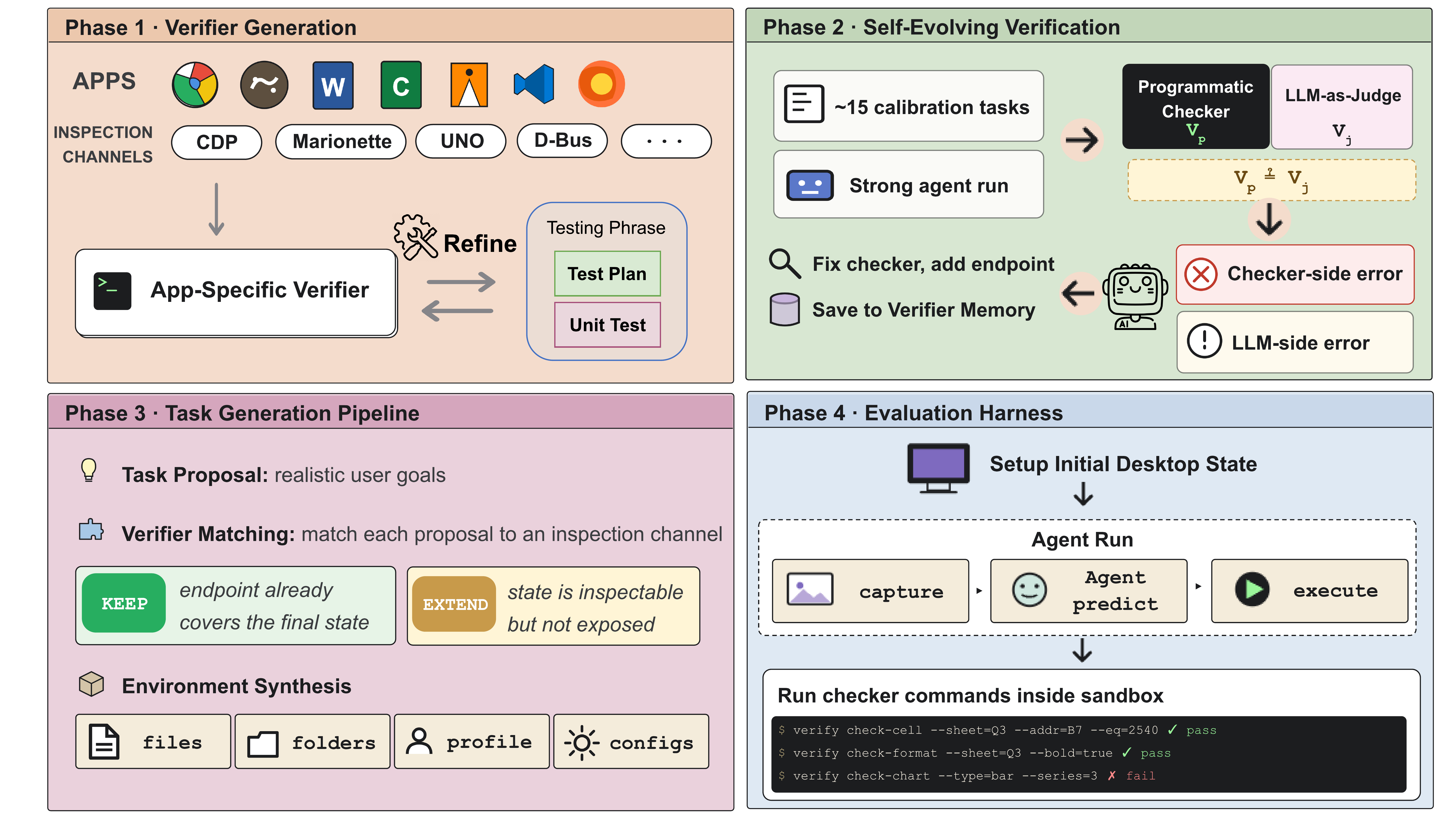}
    \caption{ Overview of the OpenComputer verifiable software-world synthesis pipeline. Phase 1 generates app-specific verifier endpoints over the most reliable inspection channels and validates it with unit and integration tests for structured, machine-checkable state. Phase 2 closes a self-evolving loop: calibration tasks drive a strong agent run, an LLM evaluator and the programmatic verifier produce verdicts that disagreement analysis attributes, and verifier memory + checker/endpoint/doc fixes refine the verifier with execution-grounded feedback. Phase 3 proposes user goals, filters by complexity and data generatability, matches against the verifier, synthesizes the environment, and emits a final task instance. Phase 4 runs the agent and computes the reward.}
    \label{fig:opencomputer-pipeline}
\end{figure*}

\section{Introduction}
Computer-use agents offer a promising path toward general-purpose AI systems that operate the same software interfaces humans use every day~\citep{agasheagent,nguyen2025gui,agashe2025agent,song2025coact}, but scaling their training and evaluation is limited by the cost of constructing realistic, reproducible desktop environments and tasks~\citep{xu2024agenttrek, he2024pc}. 

Constructing a realistic desktop task involves far more than writing a natural-language instruction. A human developer must first design a plausible user goal, then manually prepare the underlying environment state (\eg creating or editing files, configuring folders, populating spreadsheets or documents, setting browser history or bookmarks, preparing emails or calendars), and ensures that the software state is both coherent and reproducible~\cite{xie2024osworld,bonatti2024windows}. These steps are tedious, application-specific, and difficult to standardize, making large-scale task creation slow and expensive.

Beyond environment construction, computer-use tasks also require trustworthy verification of the resulting software state. In desktop settings, success is often reflected not only in visible screenshots, but also in application state, file contents, metadata, or persistent side effects~\cite{xie2024osworld,bonatti2024windows}. This makes evaluation difficult to scale: each task often requires custom inspection logic that can determine whether the intended state has actually been achieved. A natural fallback is to use an LLM-as-a-judge~\cite{liu2023g,kim2024prometheus}, but this introduces substantial limitations. LLM judgments can be sensitive to prompt wording, incomplete observations, and model-specific biases, and are often difficult to audit or reproduce across runs~\citep{wang2024large,li2025generation,thakur2025judging,zheng2023judging}. More importantly, an LLM judge may reward outcomes that appear plausible from screenshots while missing errors in the underlying software state~\citep{sumyk2026cuaaudit,cui2026agentic}. Thus, scalable synthesis for computer-use agents must be coupled with reliable inspection rather than weak proxy evaluation.

To address the dual bottlenecks of scalable environment construction and trustworthy state verification, we present \textbf{\ours}, a verifier-grounded framework for synthesizing verifiable software worlds for computer-use agents. Rather than treating verification as a downstream evaluation detail, \ours makes verification the organizing principle of environment and task construction.
It consists of four tightly coupled components as illustrated in Figure~\ref{fig:opencomputer-pipeline}. First, it builds app-specific state verifiers that undergo a strict debug-fix-retry testing loop to reliably inspect software state through stable interfaces, defining exactly which task outcomes can be checked programmatically. Second, it further improves these verifiers through an execution-grounded self-evolution loop: calibration tasks are executed in sandboxed desktops, programmatic verifier outputs are compared against criterion-level LLM judgments, and verifier-side failures are used to refine checker logic, endpoints, or documentation. Third, on top of this verifier stack, \ours synthesizes realistic user tasks through a structured pipeline that filters for difficulty, data generatability, and state inspectability. Finally, \ours provides an evaluation harness that runs agents in fresh desktop sandboxes, records full screenshot-action trajectories, and scores each run by executing verifier commands over the resulting software state.

Empirically, \ours shows that current computer-use agents still struggle to reliably complete realistic desktop tasks end to end. GPT-5.4 achieves the strongest overall performance, with a full task success rate of 68.3\%, while Claude-Sonnet-4.6 and Kimi-K2.6 reach 64.4\% and 58.8\%, respectively. Open-source agents lag substantially behind, with especially large drops relative to their reported performance on existing desktop benchmarks such as OSWorld~\cite{xie2024osworld}. Our analysis further highlights the importance of verifier-grounded benchmark construction. Hard-coded verifiers align more closely with human adjudication than an agentic LLM judge, particularly when success depends on fine-grained application state that cannot be reliably inferred from screenshots alone.

We summarize our contributions as follows:
\begin{enumerate}[leftmargin=*, itemsep=2pt]
    \item We introduce \ours, a verifier-grounded framework for synthesizing realistic software worlds for computer-use agents, where the task descriptions, environments, and verifiers for evaluation are all automatically generated without relying on manual construction.
    
    \item We empirically validate the reliability of this construction pipeline, showing that verifier-grounded evaluation aligns more closely with human adjudication than LLM-as-judge evaluation, and that the self-evolving verification layer can identify and repair verifier-side failures.

    \item We instantiate a large-scale benchmark spanning \napp desktop applications and \nexample finalized tasks, and evaluate frontier and open-source computer-use agents to show that realistic, verifier-grounded desktop workflows remain challenging for current systems.
\end{enumerate}

%% file: sections/related_work.tex
\section{Related Work}
\label{sec:related-work}
\paragraph{Benchmarks for Computer-Use Agents.}
Prior benchmarks for computer-use agents fall into two main categories: static trajectory datasets and interactive task environments. Static datasets such as
Mind2Web~\citep{deng2023mind2web} and Android in the Wild~\citep{rawles2023androidinthewild} provide broad coverage of web or mobile interfaces through human demonstrations, but primarily evaluate offline action
prediction. Interactive benchmarks more directly evaluate agents through environment feedback, including OSWorld~\citep{xie2024osworld} and Windows Agent Arena~\citep{bonatti2024windows} for desktop operating-system tasks, BEARCUBS~\citep{song2025bearcubs}, RealWebAssist~\citep{ye2026realwebassist}
for web tasks, WebArena~\citep{zhou2023webarena} and VisualWebArena~\citep{koh2024visualwebarena} for realistic web navigation, WorkArena~\citep{drouin2024workarena} and Scuba~\citep{dai2025scuba} for enterprise web workflows, and AndroidWorld~\citep{rawles2024androidworld} for mobile control.
However, these benchmarks are still largely human-curated and often limited by the number of task instances, application domains, or manually written reward checks. In contrast, \ours focuses on scaling computer-use environment construction itself.

\paragraph{Synthetic Environments for Agents.}
Recent work increasingly treats environment construction as a key bottleneck for
training interactive agents. In tool-use and function-calling settings,
AgentScaler builds simulated, database-backed API environments~\citep{fang2025towards},
Agent World Model scales code-driven multi-turn environments for RL~\citep{wang2026agent},
and Simia uses reasoning models to simulate environment feedback~\citep{li2025simulating}.
These systems demonstrate the value of scalable interactive worlds, but primarily target
abstract APIs or model-simulated feedback rather than native desktop software. Concurrent work synthesizes GUI and computer-use environments: InfiniteWeb builds
functional websites with task-centric tests~\citep{zhang2026infiniteweb}, GUI-Genesis
reconstructs mobile apps into lightweight web environments with code-native rewards~\citep{cao2026gui},
Gym-Anything~\citep{aggarwal2026gym} uses an agentic creation-and-audit loop across software applications,
and TermiGen~\citep{zhu2026termigen} and Scale-SWE~\citep{zhao2026immersion} automate executable environments for terminal and software-engineering agents. OpenComputer differs by making synthesis reward-aware from the outset: each generated desktop task is paired with verifiable reward implemented as executable checkers over inspectable application state,
rather than relying on visual proxies or LLM judgments.

%% file: sections/methodology.tex
\section{\ours}
\label{sec:methodology}
We build \ours as a verifier-grounded framework for constructing
verifiable computer-use tasks in real desktop software environments. In this section, we first define the problem setup and then describe the four key layers of \ours.

\subsection{Problem Setup}

Let $a \in \mathcal{A}$ denote a desktop application drawn from an application
set $\mathcal{A}$, and let $g \in \mathcal{G}$ denote a natural-language user
goal. Our objective is to synthesize a verifiable computer-use task instance
\[
\tau = (x, e, c)
\]
where $x$ is the task description shown to the agent, $e$ is an executable
environment initialization procedure, and $c$ is a set of machine-checkable
success criteria. Each task is executed in an initial desktop sandbox state
$s_0 \sim e$, and an agent interacts with the sandbox through screenshots and
GUI actions to produce a final state $s_T$.

The core challenge is that realistic computer-use tasks require both
environment construction and reliable verification. A goal $g$ is only useful
for benchmarking if we can: (1) materialize a coherent software world in which
the task can be performed, and (2) determine from the resulting application
state whether the goal has actually been achieved. We therefore cast environment construction as a constrained synthesis problem: given an application $a$ and a
goal $g$, generate a task instance $\tau$ such that the initial environment is
realistic, the target state is reachable through ordinary desktop interaction,
and success can be checked programmatically.

OpenComputer solves this problem through three coupled components. First, a
verifier generator
\[
\mathcal{V}(a) \rightarrow V_a
\]
builds an app-specific verifier $V_a$ that exposes structured inspection and
checking endpoints over the application state. Second, to repair residual verifier errors, a verifier-evolution procedure
\[
\mathcal{U}(V_a, D_a) \rightarrow V_a^+
\]
iteratively refines the verifier using calibration executions $D_a$ collected
from real agent runs. Third, a verifier-aware task and environment synthesis
pipeline uses the resulting verifier stack to construct task instances: given
an application $a$ and a user goal $g$, it generates an executable environment
initialization procedure
\[
\mathcal{E}(a,g,V_a^+) \rightarrow e
\]
together with a user-facing instruction $x$ and machine-checkable success
criteria $c$.

The final task synthesis pipeline combines these components to produce
benchmark instances whose environments are executable and whose rewards are
grounded in inspectable software state. The remainder of this section follows
the same order: we describe how we build app-specific verifiers, how we evolve
them from execution feedback, how we generate verifier-grounded task
environments, and how we evaluate agents with structured reward computation.

\subsection{Verification Stack}
\label{sec:verification-stack}
Verification is central to OpenComputer because realistic desktop tasks are only
useful for training or evaluation when their outcomes can be checked reliably. Many success conditions are hidden in application state rather than visible in screenshots. The
verification stack therefore defines what can be trusted as reward, and ensures that task generation and evaluation are grounded in reproducible, machine-checkable evidence.

\subsubsection{Verifier Generation}
\label{sec:verifier-generation}

\begin{wrapfigure}{r}{0.47\columnwidth}
    \centering
    \vspace{-20pt}
    \includegraphics[width=0.47\columnwidth]{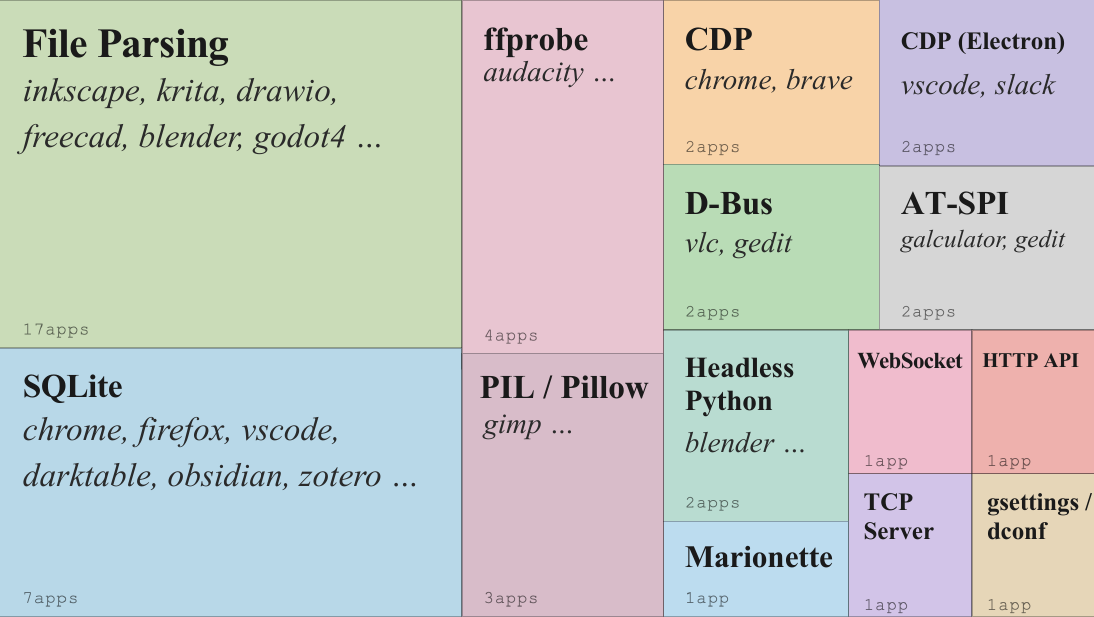}
    \vspace{-12pt}
    \caption{Example application endpoint specification used by OpenComputer verifiers.} 
    \vspace{-23pt}
    \label{fig:app-endpoint-example}
\end{wrapfigure}

Each supported application in the environment is paired with a
synthetic Python verifier module that runs inside the sandbox and exposes a set
of CLI subcommands with JSON outputs. These verifiers serve as stable
inspection interfaces for downstream task generation and evaluation. Rather
than focusing only on an application's primary document content, they are
designed to cover all reliably inspectable state surfaces available for that
application, including content state, preferences, plugins, history, bookmarks, file I/O, project structure, media state,
graphical attributes, and metadata. In the notation of
Section~\ref{sec:methodology}, for each application $a \in \mathcal{A}$ we
instantiate an app-specific verifier $V_a = \mathcal{V}(a)$.

\paragraph{Inspection channels.} To achieve this coverage, verifier endpoints query the most reliable
application-specific inspection channels available in the sandbox. Depending on
the target application, these channels may include browser debugging protocols,
D-Bus, LibreOffice UNO, SQLite-backed profile databases, accessibility state,
or direct parsing of saved files as shown in Figure~\ref{fig:app-endpoint-example}. In this way, verification is grounded in the
actual observable state of the application rather than in heuristic matching or
surface-level script checks.

\paragraph{Endpoint construction.} Verifier development follows a fixed pipeline. The agent first enumerate the inspectable state surfaces of the target application and map each surface to a
concrete verification channel. For example, browser-oriented tasks can often be
verified through remote debugging APIs, office tasks through UNO interfaces or
document parsing, and configuration-oriented tasks through SQLite databases. Based on this mapping, the agent implement query endpoints and
\texttt{check-*} endpoints that expose these states as structured JSON, and then
document them in an application-specific README so that later pipeline stages
can treat the verifier as a well-defined interface.

\paragraph{Verifier testing protocol.} The agent treat verifiers as software artifacts rather than ad hoc scripts. Each verifier includes an endpoint reference, a written test plan, and live integration tests against the real sandboxed application. The test plan covers expected assertions, realistic fixtures, positive and negative cases, JSON-validity checks, and common failure modes such as missing arguments, nonexistent paths, or inactive applications. For document-centric applications, the agent generate rich synthetic artifacts with realistic structure rather than toy files. Failed endpoints enter a debug-fix-retry loop until they become reliable, since unstable verifiers can produce misleading rewards.

\subsubsection{Self-Evolving Verification Layer}
\label{sec:self-evolving-verification}
After the initial verifier for an application is generated and passes its
unit and integration tests, we further refine it through a
self-evolving verification layer. The goal of this layer is to expose
residual verifier issues that may not appear in synthetic tests alone, such
as brittle assumptions about application schemas, incomplete endpoint
coverage, or mismatches between documented and actual software behavior.

\paragraph{Calibration executions.} For each application, we generate a small calibration set of approximately
15 easy-to-medium tasks that are expected to be solvable by a
state-of-the-art computer-use agent. These tasks are not used to benchmark
agent performance. Instead, they serve as execution-grounded probes for
stress-testing the verifier before it is used for large-scale task synthesis
and evaluation. We run the selected agent in a persistent desktop sandbox,
record the full trajectory, and cache the resulting final environment state.
The resulting execution can be viewed as taking the sandbox from an initialized
state $s_0 \sim e$ to a realized terminal state $s_T$, and this recorded run is
then treated as fixed throughout the refinement procedure.

\paragraph{Disagreement diagnosis.} Given each fixed execution, an LLM evaluator inspects the trajectory,
post-action observations, and final state to produce a criterion-level
reference verdict. Independently, the programmatic verifier is executed
against the same final state to produce a structured machine verdict. A
comparator aligns the two verdicts criterion by criterion and identifies
disagreements. Disagreements attributed to genuine agent failures are
discarded, while disagreements attributed to verifier-side errors are used as
feedback for improving the verifier implementation, endpoint documentation,
or task-checking logic.

\paragraph{Bounded verifier refinement.} The verifier evolution step is restricted to the verification stack:
it may modify checker code, endpoint implementations, or verifier documentation,
but does not alter the cached trajectory, sandbox state, task objective, or expected outcome.
The revised verifier is re-executed on the same cached final state, and the process iterates
until the updated verifier $V_a^+ = \mathcal{U}(V_a, D_a)$ agrees with the reference judgment
on verifier-attributed criteria, or until a fixed evolution budget is exhausted.
When verifier-side issues are repaired, OpenComputer records the failed assumption and
corrective action as an app-specific lesson that can be reused during future verifier extension
and task generation.

This layer provides an additional feedback channel between real software
execution and verifier construction. By running strong agents on simple and
moderate calibration tasks, OpenComputer can identify which endpoints are underspecified, and
which verifier assumptions fail under realistic interaction. A concrete example of this stage is shown in Appendix~\ref{app:self-evolving-verification-case-study}.

\subsection{Task Generation Pipeline}

Tasks are generated through a verifier-aware synthesis process that balances realism, difficulty, and checkability. The generator first proposes candidate tasks from the perspective of realistic user goals, without directly conditioning on the available verifier endpoints. This encourages task diversity and avoids overfitting the benchmark to what is already easy to check. Candidate tasks are then filtered for complexity and data generatability: we prioritize multi-step workflows in the upper half of the difficulty scale and reject tasks that are too short, overly linear, trivial, or difficult to instantiate with coherent input artifacts.

Accepted proposals are then grounded in the verification stack. If the intended state can be checked by an existing endpoint, the task is retained directly. If the outcome is inspectable but not yet exposed, the verifier is extended with a new endpoint following the verifier-generation procedure in Section~\ref{sec:verifier-generation}. Finally, the system materializes each task by generating and packaging the required files, folders, profiles, configurations, or other input artifacts. Each finalized task is stored as a \texttt{task.json} instance $\tau=(x,e,c)$, where $x$ is the user-facing instruction, $e$ initializes the sandbox, and $c$ specifies the executable success criteria. This process turns open-ended desktop workflows into reproducible benchmark instances with machine-checkable rewards.

To prevent coverage collapse, the task generator includes a task-extension workflow. We periodically review each application's task set by feature area, identify missing or repetitive workflows, and prioritize gaps with reliable verification paths. New candidate tasks for these gaps are then passed through the same four-stage proposal, filtering, verification, and environment-synthesis pipeline.

\subsection{Evaluation Harness and Reward Computation}

At evaluation time, the harness uploads the verifier and task artifacts into a fresh sandbox, launches the target application, and runs a screenshot-action
loop with the chosen agent. At each step, the system captures the current desktop framebuffer, feeds it to the agent, executes the predicted action, and
logs the resulting reasoning, action sequence, and screenshot.  In the formalization above, the evaluation harness executes the task instance $\tau = (x, e, c)$ by first sampling $s_0 \sim e$ and then checking whether the evaluated agent's interaction trajectory reaches a terminal state $s_T$ that satisfies
$c$.

After the agent stops or reaches a step budget, the harness attempts a final
save action for applications where persistence matters. Verification is then
performed by executing the task's checker commands inside the sandbox. The task reward is the fraction of checks that pass, $R = N_{\mathrm{pass}} / N_{\mathrm{total}}$. This scoring scheme supports partial credit while preserving exact, machine-checkable success conditions. As an optional quality-control step, we randomly apply the self-evolving verification procedure from Section~\ref{sec:self-evolving-verification} to update checkers on finalized tasks.

\input{figure/data_statistics}
\subsection{\ours Release}
We release \ours as an extensible infrastructure for both training and evaluating computer-use agents in verifiable software environments. The release includes 33 desktop applications and 1,000 finalized tasks, together with app-specific verifier modules, task specifications, environment-initialization
scripts, and an execution harness. Summary statistics of the released synthetic benchmark are reported in Table~\ref{tab:data_statistics}. \ours supports both local and cloud-scale execution. Users can run tasks locally with Docker-based sandboxes, deploy the same stack on self-hosted or cloud machines such as AWS, Tencent Cloud, or E2B for parallel rollouts. Beyond fixed evaluation, OpenComputer also naturally supports extension to training pipelines: future researchers can collect trajectories, filter successful or partially successful runs, build SFT data, and use machine-checkable rewards for RL or rejection sampling. Users can also extend existing applications or add new ones through the same verifier-guided task and environment synthesis workflow.

%% file: figure/data_statistics.tex
\begin{table}[t]
\centering
\small
\caption{Summary statistics of the OpenComputer benchmark.}
\begin{tabular}{c c c c c}
\toprule
\textbf{Applications} &
\textbf{Tasks} &
\textbf{Avg. Verifier Endpoints / App} &
\textbf{Avg. Checks / Task} &
\textbf{Avg. Seed Files / Task} \\
\midrule
33 & 1000 & 17.7 & 6.9 & 1.3 \\
\bottomrule
\end{tabular}
\label{tab:data_statistics}
\end{table}

%% file: sections/experiment.tex
\section{Experiment}
\label{sec:experiment}
\subsection{Experimental Setup}
\label{sec:experimental-setup}
We design our experiments to evaluate whether OpenComputer provides reliable and challenging
software-world environments for computer-use agents. Our evaluation focuses on two questions:
(1) whether current frontier and open-source agents can complete the synthesized tasks, and
(2) whether our verifier-grounded reward computation can measure both exact success and
partial progress across heterogeneous desktop applications.
\input{figure/benchmark_result}
\paragraph{Benchmark.}
We evaluate agents on the finalized OpenComputer task suite. Each task consists of a natural-language
instruction, an executable sandbox initialization, and a set of machine-checkable success criteria.
The benchmark spans 33 desktop applications. For each task, the agent is placed in a fresh desktop sandbox initialized with the required files, profiles, configuration state, and application artifacts. The agent then interacts with the live GUI
through screenshots and desktop actions until it stops or reaches the step budget. We include OSWorld-Verified~\citep{xie2024osworld} as an external reference to contextualize model performance against a widely used desktop-agent benchmark.

\paragraph{Models.}
We evaluate a mixture of frontier proprietary agents and open-source computer-use models. The
main models include GPT-5.4~\citep{openai2026gpt54}, Claude-Sonnet-4.6~\citep{anthropic2026sonnet46}, Kimi-K2.6~\citep{moonshot2026kimi26}, Gemini-3-Flash~\citep{google2025gemini3flash}, Qwen-3.5-27B~\citep{qwen2026qwen35},
Qwen-3.5-9B~\citep{qwen2026qwen35}, EvoCUA-8B~\citep{xue2026evocua}, and GUI-OWL-1.5-8B~\citep{xu2026mobile}. For Gemini-3-Flash, which does not
provide a built-in desktop action space in our evaluation setting, we prompt the model to emit actions in a Qwen-style computer-use format. All open-source models except Kimi-K2.6 (which we use the official APIs) are deployed with two H100 GPUs.

\paragraph{Metrics.}
We report both task-level and criterion-level metrics. The primary task-level metric is \emph{success
rate}, defined as the fraction of tasks for which all required criteria are satisfied. Because many desktop
tasks contain multiple independent requirements, we also report \emph{average reward}, defined as
the mean fraction of passed verifier checks, this metric gives partial credit when an agent completes some but not all required subtasks. To measure
efficiency, we additionally report the average number of interaction steps and the average wall-clock
time per step.

\subsection{Main Results Analysis}

Table~\ref{tab:model_results} reports the overall performance and efficiency of representative 
computer-use agents on OpenComputer. The results show that OpenComputer is challenging even
for the strongest current agents. GPT-5.4 achieves the best overall performance, with an average
reward of 88.4\% and a task success rate of 68.3\%, but it still fails to completely solve nearly one
third of the benchmark tasks. Claude-Sonnet-4.6 and Kimi-K2.6 follow closely, reaching success
rates of 64.4\% and 58.8\%, respectively. This indicates that frontier models can often make
substantial partial progress, but reliable end-to-end task completion in realistic desktop software
remains far from saturated.

GPT-5.4 is also the most efficient agent in terms of interaction length. It completes tasks in only
19.0 steps on average, substantially fewer than Claude-Sonnet-4.6, Kimi-K2.6, and the open-source
models. One reason is that GPT-5.4 frequently combines multiple low-level operations into a single
computer-control step, reducing the number of interaction rounds needed to complete a task. In
addition, GPT-5.4 does not emit long reasoning traces in our evaluation setting but only the executable actions, which reduces output overhead and improves per-step execution efficiency. This combination of shorter trajectories
and lower textual overhead makes it particularly effective for controlling computer environments.

The external OSWorld-Verified scores provide additional context. Several open-source models have moderate reported OSWorld performance, but their success rates drop substantially on OpenComputer.
For example, GUI-OWL-1.5-8B has a reported OSWorld score of 52.3\%, but achieves only 5.7\%
success on our benchmark; EvoCUA-8B similarly drops from 46.1\% on OSWorld to 10.9\% on
\ours. This gap suggests that these models have limited cross-benchmark generalization,
and that strong performance on existing desktop benchmarks does not necessarily transfer to the
broader and more heterogeneous software settings covered by \ours.

%% file: figure/benchmark_result.tex
\begin{table}[t]
\centering
\small
\setlength{\tabcolsep}{5pt}
\renewcommand{\arraystretch}{1.12}
\caption{Performance and efficiency comparison across computer-use agents on our benchmark, with OSWorld-Verified reported as an external reference when available. The OSWorld column summarizes the publicly reported OSWorld-Verified score for the corresponding model. Success rate reports the fraction of tasks completed successfully. Average steps and time (seconds) per step capture interaction efficiency. Average reward measures the mean checklist-based score over all tasks.}
\vspace{-0.5em}
\begin{tabular}{l|r|rrrr}
\toprule
\textbf{Model} 
& \textbf{OSWorld} 
& \textbf{Success Rate} 
& \textbf{Avg. Steps} 
& \textbf{Time/Step} 
& \textbf{Avg. Reward} \\
\midrule
GPT-5.4           & 75.0\% & 68.3\% & 19.0 & 16.5 s  & 88.4\% \\
Claude-Sonnet-4.6 & 72.5\% & 64.4\% & 31.5 & 20.8 s  & 76.6\% \\
Kimi-K2.6         & 73.1\% & 58.8\% & 35.7 & 33.0 s  & 70.7\% \\
Qwen-3.5-27B      & 56.2\% & 32.3\% & 33.1 & 57.3 s  & 59.4\% \\
Gemini-3-Flash    & --     & 16.4\% & 25.4 & 9.0 s   & 37.0\% \\
EvoCUA-8B         & 46.1\% & 10.9\% & 67.0 & 9.7 s   & 38.1\% \\
Qwen-3.5-9B       & 41.8\% & 7.8\%  & 39.3 & 17.8 s  & 31.7\% \\
GUI-OWL-1.5-8B    & 52.3\% & 5.7\%  & 73.6 & 9.43 s  & 27.8\% \\
\bottomrule
\end{tabular}
\label{tab:model_results}
\end{table}

%% file: sections/analysis.tex
\section{Analysis}
\label{sec:analysis}
\subsection{Agentic LLM-as-Judge vs. Hard-Coded Verification}
\label{sec:llm-as-judge-comparison}
We use LLM-as-judge as a two-stage agentic pipeline. The judge first reads the
reasoning and action trace to identify a small set of steps that are most
likely to contain evidence for each criterion. It then scores each criterion
from these steps' corresponding screenshots, with the option to retrieve more screenshots when existing ones are not sufficient. This setup makes long trajectories tractable to
inspect and is useful for diagnosing failures during task synthesis.
To quantify the gap between these two evaluation strategies, we sample 120
tasks and send the same completed trajectories to human annotators. We then
score the 120 trajectories with two automated evaluators: an LLM judge and our
final hard-coded verifier. We use the same per-item checklist for both methods.
For each task, the item-level decisions are aggregated into a task-level
verdict, and we compare that verdict against the human label. 

\input{figure/compare_with_llm_as_judge.tex}
Figure~\ref{fig:judge-vs-checker-human-alignment}
shows that the hard-coded verifier aligns much better with human judgment at
both levels: it matches human verdicts on 113 out of 120 tasks, whereas the LLM
judge reaches 95 out of 120, and it also achieves higher per-item checklist
agreement with human annotations (97.3\% versus 92.2\%).
In dense
desktop interfaces, semantically important mistakes are often visually tiny: a
model may type two tokens into one spreadsheet cell instead of two adjacent
cells, apply a formatting change to the wrong selection, or edit a field inside a collapsed panel that is only partially visible. These runs can look approximately correct from pixels alone. A hard-coded verifier instead reads the exact application state and can thus distinguish near-miss visual outputs from true task completion.

The gap is even larger for applications with heavy terminal usage or agents with mixed action spaces. In environments such as Blender or developer tools, success often depends on scrollback logs or intermediate artifacts that are not simultaneously visible on screen. An LLM judge only sees a narrow window of
the terminal and must infer the rest from partial evidence, while a programmatic verifier can directly inspect post-execution application state. Appendix~\ref{app:llm-as-judge-problem} presents two concrete visual examples of these failure modes.


\subsection{Comparing GUI Agents with CLI Agents}
Because \ours verifies final application state rather than a particular interaction trace, it can in principle evaluate agents that reach the same target state through different control interfaces. We therefore compare GUI and CLI agents on a shared CLI-compatible subset to test whether \ours’s verifier-grounded tasks transfer beyond screenshot-action GUI control, and to quantify the trade-off between visual grounding and programmatic execution efficiency.
Since many OpenComputer applications are inherently GUI-centric and cannot be meaningfully executed from a terminal alone, we
construct a controlled subset by removing applications whose tasks are not
suitable for CLI execution. This leaves 14 applications and 343 tasks that are
compatible with both GUI and CLI settings. For the CLI setting, we use Claude
Sonnet 4.6 with Claude Code, where the agent can combine CLI-Anything skills~\citep{hkuds2026clianything},
Bash commands, and Python scripts to inspect files, manipulate artifacts, and
execute application-specific operations.

\input{figure/gui_and_cli}
Table~\ref{tab:gui_cli_overall} shows the comparison. On this shared subset,
GUI agents still achieve higher pass rates than the CLI agent. This suggests that even when tasks are selected to be CLI-solvable, visual interaction provides useful grounding for many desktop workflows. At the same time, the CLI agent is substantially faster. Claude Code completes
tasks in 141 seconds on average, compared with 288 seconds for GPT-5.4 GUI
control and 622 seconds for Claude Sonnet 4.6 GUI control. This reflects the efficiency advantage of command-line
execution: the agent can bypass slow screenshot-action interaction loops and directly manipulate files, run scripts, or invoke application-level tools.
\subsection{Ablation: Self-Evolving Verification}
\input{figure/ablation_of_self_evolving.tex}
We ablate the contribution of the self-evolving verification layer by measuring how often it can
identify and repair checker-side errors. We generate 450 simple calibration tasks and run the
self-evolution procedure with a maximum repair budget of three iterations per task. These tasks are
used only to probe verifier reliability, not to measure agent capability.

Among the
450 calibration executions, 159 tasks exhibit at least one disagreement between the programmatic
checker and the reference evaluation. After categorizing the disagreement source, we find that
76 cases are attributable to checker-side errors rather than agent failures.
The self-evolution procedure repairs 68 of these 76 checker-side cases, corresponding to an 89.4\%
repair rate.

Table~\ref{tab:self_evolution_rounds} further breaks down the repair process. Most checker-side
errors are fixed quickly: 47 cases are repaired after one iteration, 15 after two iterations, and only
6 require the full three-iteration budget. The remaining 8 cases are not resolved within the budget.
We compare the pre- and post-evolution checkers on the
same 120-task human-annotated comparison set used in
Section~\ref{sec:llm-as-judge-comparison}. As a result, human-checker agreement improves from 85.2\% before self-evolution to 94.1\% after
self-evolution. This suggests that the self-evolving layer provides a useful debugging signal for
improving verifier reliability while preserving programmatic, auditable evaluation.

%% file: figure/compare_with_llm_as_judge.tex
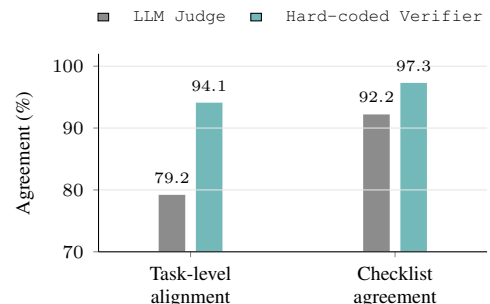
\begin{wrapfigure}{r}{0.42\columnwidth}
    \centering
    \vspace{-10pt}
    \begin{tikzpicture}
    \colorlet{judgeColor}{black!45}
    \colorlet{verifierColor}{teal!55}

    \def\figAxisFont{\fontsize{7.5pt}{8.5pt}\selectfont}
    \def\figSmallFont{\fontsize{6.5pt}{7.5pt}\selectfont}

    \begin{axis}[
        width=\linewidth,
        height=4.2cm,
        ybar,
        bar width=10pt,
        ymin=70, ymax=102,
        xmin=-0.45, xmax=1.45,
        xtick={0,1},
        xticklabels={{Task-level\\alignment},{Checklist\\agreement}},
        xticklabel style={font=\figAxisFont, align=center},
        ytick={70,80,90,100},
        yticklabel style={font=\figAxisFont},
        ylabel={Agreement (\%)},
        ylabel style={font=\figAxisFont},
        axis lines=left,
        axis line style={-},
        axis on top,
        tick align=outside,
        major tick length=2pt,
        ymajorgrids=true,
        grid style={gray!20},
        nodes near coords,
        point meta=y,
        every node near coord/.append style={
            /pgf/number format/fixed,
            /pgf/number format/precision=1,
            font=\figSmallFont,
            yshift=1pt,
        },
        legend style={
            draw=none,
            font=\figSmallFont,
            at={(0.5,1.09)},
            anchor=south,
            legend columns=2,
            column sep=0.7em,
        },
        legend image code/.code={
            \draw[#1] (0cm,-0.05cm) rectangle (0.14cm,0.05cm);
        },
        clip=false,
    ]

    \addplot[
        draw=none,
        fill=judgeColor,
        bar shift=-7pt,
    ] coordinates {
        (0,79.2)
        (1,92.2)
    };

    \addplot[
        draw=none,
        fill=verifierColor,
        bar shift=7pt,
    ] coordinates {
        (0,94.1)
        (1,97.3)
    };

    \legend{\texttt{LLM Judge}, \texttt{Hard-coded Verifier}}

    \end{axis}
    \end{tikzpicture}
    \vspace{-1.3em}
    \captionof{figure}{Alignment with human adjudication on a 120-task comparison set.}
    \label{fig:judge-vs-checker-human-alignment}
\end{wrapfigure}

%% file: figure/gui_and_cli.tex
\begin{wraptable}{r}{0.45\columnwidth}
\centering
\setlength{\tabcolsep}{3pt}
\caption{Overall GUI\textendash CLI pass-rate and execution-time (per task) comparison. For the CLI Agent, we use Claude Code (v2.1.129).}
\label{tab:gui_cli_overall}
\vspace{-0.5em}

\resizebox{\linewidth}{!}{%
\begin{tabular}{@{}llcc@{}}
\toprule
\textbf{Setting} & \textbf{Model} & \textbf{Success Rate (\%)} & \textbf{Time (s)} \\
\midrule
GUI & GPT-5.4 & 75.2 & 288 \\
GUI & Claude Sonnet 4.6 & 73.0 & 622 \\
\midrule
CLI & Claude Sonnet 4.6 & 67.2 & 141 \\
\bottomrule
\end{tabular}%
}

\end{wraptable}

%% file: figure/ablation_of_self_evolving.tex
\begin{wraptable}{r}{0.38\columnwidth}
\vspace{-1.3em}
\centering
\small
\setlength{\tabcolsep}{3pt}
\caption{Repair efficiency and human-checker agreement improvement from self-evolving verification.}
\vspace{-0.5em}
\label{tab:self_evolution_rounds}
\begin{tabular}{lr}
\toprule
Metric & Value \\
\midrule
Fixed in 1 round & 47 \\
Fixed in 2 rounds & 15 \\
Fixed in 3 rounds & 6 \\
Not fixed within budget & 8 \\
\midrule
Agreement before evolution & 85.2\% \\
Agreement after evolution & 94.1\% (+8.9\%) \\
\bottomrule
\end{tabular}
\vspace{-1.0em}
\end{wraptable}

%% file: sections/conclusion.tex
\section{Conclusion}
\label{sec:conclusion}
We introduced \textbf{\ours}, a verifier-grounded framework for building verifiable software worlds for computer-use agents. \ours makes inspectable application state a core design constraint across verifier construction, task synthesis, and benchmark execution. This enables the automatic generation of executable desktop tasks with machine-checkable success criteria while preserving the diversity and realism of real software workflows. In its current form, \ours covers 33 desktop applications and 1,000 finalized tasks across browsers, office tools, creative software, development environments, communication tools, and system utilities.

We show that realistic desktop environments expose important failure modes in current computer-use agents. Frontier agents often make meaningful partial progress, but reliable end-to-end completion remains difficult when success depends on fine-grained application state, persistent files, metadata, or hidden side effects. Our comparison between LLM-as-judge and hard-coded verification further shows that screenshot-based or trajectory-level judgments can miss subtle but consequential errors, whereas executable verifiers can directly inspect the final software state.

More broadly, we view \ours as infrastructure for scaling computer-use research. Progress requires not only stronger models, but also trustworthy environments, grounded rewards, reproducible task construction pipelines, and verifiers that support both evaluation and training. By coupling realistic software worlds with machine-checkable feedback, \ours provides a foundation for studying agent reliability, collecting grounded trajectories, analyzing failures, and improving agents through supervised learning, rejection sampling, or reinforcement learning. We hope this work and the released repository help make future computer-use systems more reliable, measurable, and aligned with real software outcomes.

%% file: Appendix/limitations.tex
\section*{Limitations and Future Work}
\label{app:limitations}

Although OpenComputer is designed around executable, hard-coded verification, not every realistic desktop task can be fully reduced to reliable programmatic checks. Some generated tasks require visual or geometric judgments that are difficult to express using application state alone. For example, in Draw.io, a verifier can often inspect the existence of shapes, labels, and connector objects, but it may be difficult to determine with high confidence whether an arrow visually and semantically connects two specific boxes in the intended way without inspecting the rendered screenshot. Similar cases arise when the desired outcome depends on spatial layout, visual alignment, or other presentation-level properties that are only partially exposed through file formats or application APIs.

When a generated task contains criteria that cannot be reliably checked by a hard-coded verifier, we mark those criteria as requiring LLM-based visual judgment rather than treating them as fully programmatic rewards. However, to keep the official benchmark auditable and reproducible, we exclude such tasks from the main benchmark and from all reported evaluation results. In the current task-generation process, we found 17 generated tasks with at least one success criterion that could not be fully verified by hard-coded checkers; these tasks were retained only for diagnostic analysis and were not included in the finalized OpenComputer benchmark. We will release these tasks in the repository, together with the procedure used to identify visually grounded criteria and the LLM-as-judge pipeline used for analysis. This provides a controlled starting point for future work on hybrid verification, where executable state checks can be combined with visual judgments for desktop tasks whose success depends on layout, geometry, or rendered appearance.

%% file: Appendix/repair_module.tex
\section{Case Study: Self-Evolving Verification in a Programmatic Verifier}
\label{app:self-evolving-verification-case-study}

This appendix provides a concrete example of the self-evolving verification
layer described in Section~\ref{sec:self-evolving-verification}. The goal of this
layer is to use execution-grounded feedback to refine the verification stack and
improve future task synthesis. The example below illustrates a common failure
mode: the agent completed the task, but the verifier queried an outdated
application schema and therefore incorrectly marked several satisfied criteria as
failed. By comparing the programmatic verdict against an LLM reference judgment
on a fixed trajectory, the system identifies the verifier-side defect and updates
the verification logic accordingly.

\paragraph{Task.}
We use the darktable task \texttt{darktable\_batch\_rate\_and\_tag}. The agent is
instructed to import three images, create the tag \texttt{batch\_processed},
attach the tag to all three images, and assign ratings of one, three, and five
stars to \texttt{img\_001.png}, \texttt{img\_002.png}, and
\texttt{img\_003.png}, respectively. The recorded run was produced by
\texttt{kimi-k2.6} and completed in 53 interaction steps. The trajectory was then
frozen and reused throughout the self-evolution procedure.

\paragraph{Reference judgment.}
An LLM judge inspected the full trajectory, post-action screenshots, and final
state. It judged all ten criteria as satisfied: the three images were imported,
the tag \texttt{batch\_processed} was visible and attached to all images, and the
final star ratings matched the requested values. In particular, the judge found
that the tag was created while all three images were selected, and that the final
state showed the expected image information and rating flags for each image.

\begin{table}[H]
\centering
\small
\begin{tabular}{llccc}
\toprule
Round & Source & Passed & Failed & Divergences \\
\midrule
0 & Programmatic verifier before evolution & 6 & 4 & 4 \\
0 & LLM reference judgment & 10 & 0 & -- \\
1 & Programmatic verifier after evolution & 10 & 0 & 0 \\
\bottomrule
\end{tabular}
\caption{
Self-evolution outcome for \texttt{darktable\_batch\_rate\_and\_tag}. The initial
verifier incorrectly failed four tag-related criteria. After updating the
verification logic using execution-grounded feedback, the programmatic verdict
agreed with the LLM reference on all ten criteria.
}
\label{tab:darktable-self-evolving-outcome}
\end{table}

\paragraph{Detected disagreement.}
The comparator found four disagreements between the LLM reference judgment and
the programmatic verifier. All four involved tag state: whether the tag
\texttt{batch\_processed} existed, and whether it was attached to each of the
three imported images. In each case, the LLM judge returned \textsc{True}, while
the verifier returned \textsc{False}. The comparator classified all four
disagreements as verifier-side failures, meaning that the task had been completed
but the checker misjudged the final state.

\begin{table}[H]
\centering
\small
\caption{
Criterion-level disagreements before self-evolution. All failures share the
same root cause: the verifier checked tag metadata using an outdated database
assumption.
}
\begin{tabular}{clccc}
\toprule
Criterion & Description & Verifier & Judge & Classification \\
\midrule
3 & Tag \texttt{batch\_processed} exists & False & True & \texttt{verifier\_wrong} \\
4 & \texttt{img\_001} has tag & False & True & \texttt{verifier\_wrong} \\
5 & \texttt{img\_002} has tag & False & True & \texttt{verifier\_wrong} \\
6 & \texttt{img\_003} has tag & False & True & \texttt{verifier\_wrong} \\
\bottomrule
\end{tabular}
\label{tab:darktable-self-evolving-disagreements}
\end{table}

\paragraph{Root cause.}
All four failures were caused by a single verifier bug. The darktable verifier
assumed that the table \texttt{tags} lived in \texttt{library.db}. In the current
darktable state, however, tag definitions are stored in \texttt{data.db}, while
image-tag associations remain in \texttt{library.db}. As a result, tag-related
SQL queries failed with a missing-table error and were counted as negative
verifier results, even though the final application state contained the expected
tag assignments.

\paragraph{Verifier evolution.}
The self-evolving layer was allowed to modify only the verifier implementation
and documentation, not the agent trajectory, sandbox state, task specification,
or expected output. The verifier update made three changes. First,
\texttt{check\_tag\_exists} and the corresponding tag-query endpoint were
rerouted to query \texttt{data.db}. Second, the image-tag checker was rewritten
to join \texttt{library.db}'s \texttt{tagged\_images} table with
\texttt{data.db}'s \texttt{tags} table. Third, the verifier documentation was
updated to reflect the actual darktable schema. These changes preserve the same
public checker interface while aligning the internal inspection logic with the
real application state.

\begin{table}[H]
\centering
\small
\caption{
Summary of the verifier evolution. The public checker interface was unchanged;
only the internal SQL source and join path were updated.
}
\begin{tabular}{p{0.28\linewidth}p{0.31\linewidth}p{0.31\linewidth}}
\toprule
Check & Before evolution & After evolution \\
\midrule
Tag existence &
Query \texttt{main.tags} inside \texttt{library.db}. &
Query \texttt{main.tags} inside \texttt{data.db}. \\
\addlinespace
Image-tag assignment &
Join \texttt{main.tagged\_images} with \texttt{main.tags} inside \texttt{library.db}. &
Join \texttt{main.tagged\_images} from \texttt{library.db} with \texttt{data.tags} from attached \texttt{data.db}. \\
\bottomrule
\end{tabular}
\label{tab:darktable-self-evolving-update}
\end{table}

\paragraph{Before and after.}
The tag-existence check required only changing the database queried by the
existing SQL statement:
\begin{verbatim}
# Before
rows = _query_sqlite(LIBRARY_DB, sql, (tag_name, f"%|{tag_name}"))

# After
rows = _query_sqlite(DATA_DB, sql, (tag_name, f"%|{tag_name}"))
\end{verbatim}

The image-tag checker required a cross-database join:
\begin{verbatim}
# Before
SELECT t.id AS tag_id, t.name AS tag_name
FROM main.tagged_images ti
JOIN main.tags t ON ti.tagid = t.id
WHERE ti.imgid = ? AND (t.name = ? OR t.name LIKE ?)

# After
SELECT t.id AS tag_id, t.name AS tag_name
FROM main.tagged_images ti
JOIN data.tags t ON ti.tagid = t.id
WHERE ti.imgid = ? AND (t.name = ? OR t.name LIKE ?)
\end{verbatim}

\paragraph{Outcome.}
After self-evolution, the verifier was re-executed on the same cached final
state. The updated verifier passed all ten criteria and had zero remaining
divergences from the LLM reference judgment. This example demonstrates how the
self-evolving verification layer provides an additional feedback channel for the
synthesis pipeline: it identifies brittle verifier assumptions, such as
application schema drift, updates the executable inspection logic, and records
which application states require more careful grounding in future task
generation. In this way, OpenComputer improves its verifier stack over time
while preserving the core principle that agent performance is scored by
executable, application-grounded checks.

%% file: Appendix/llm_as_judge_problem.tex
\section{Case Study: Comparison between LLM as Judge and Hard-Coded Verifier}
\label{app:llm-as-judge-problem}

This appendix illustrates why we use LLM-as-judge only as a reference signal for
verifier debugging, rather than as the final benchmark reward.

\begin{figure}[t]
    \centering
    \includegraphics[width=0.98\linewidth]{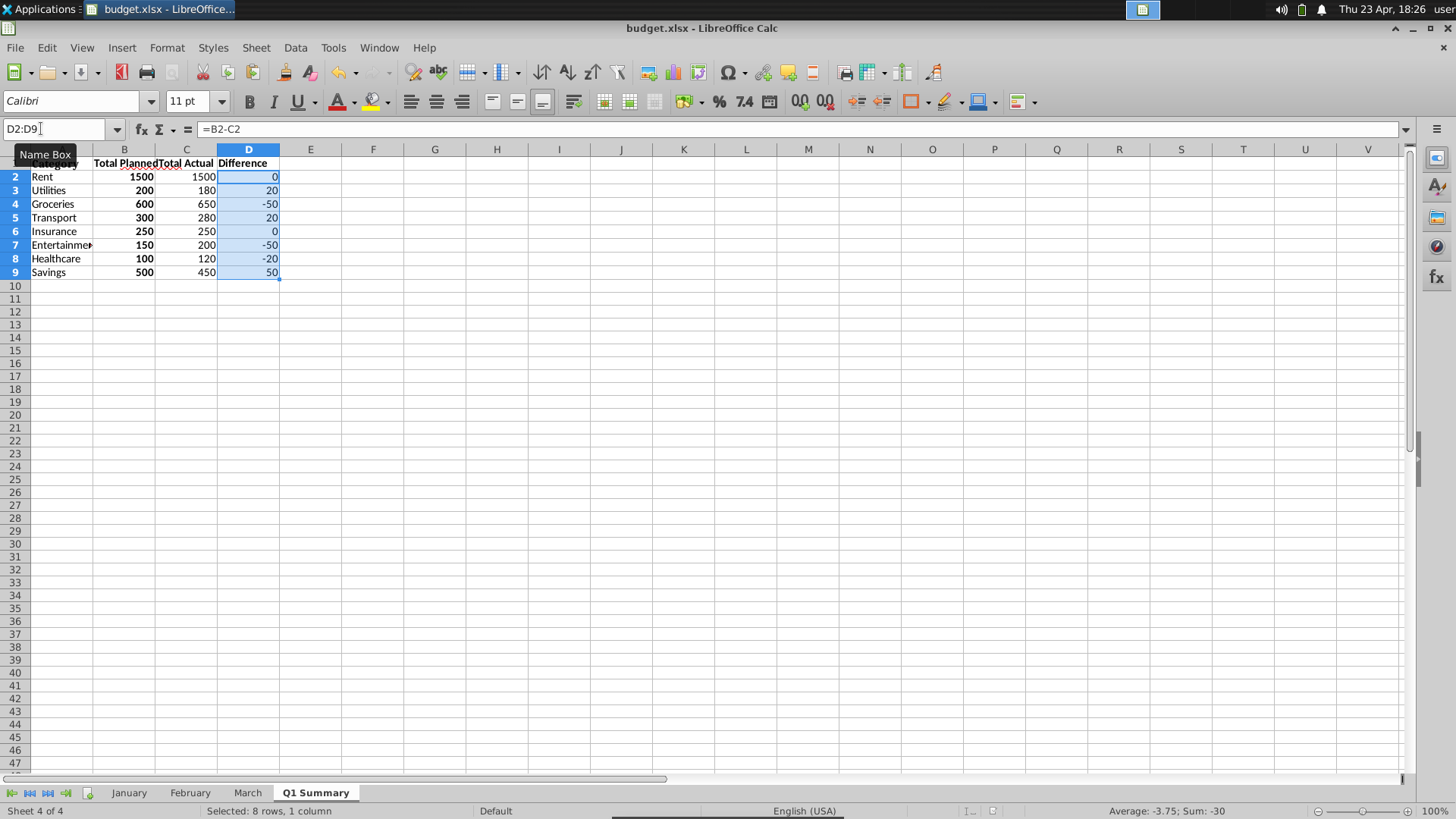}
    \caption{A dense spreadsheet-style interface where the visual output looks
    almost correct, but the state is wrong: the agent typed one long token into a
    single cell instead of filling two adjacent cells. This is difficult to judge
    reliably from pixels alone, but trivial to detect from the underlying
    workbook state.}
    \label{fig:llm-judge-dense-ui}
\end{figure}

\paragraph{Failure mode 1: dense interfaces hide exact state.}
In spreadsheet-like applications, the difference between success and failure may
be encoded in a single cell boundary, a hidden formula, or a small formatting
change. Figure~\ref{fig:llm-judge-dense-ui} shows a representative example in
which the agent types \texttt{alpha beta} into one cell, although the task
requires \texttt{alpha} and \texttt{beta} to be entered into two adjacent cells.
To a screenshot-based judge, the rendered sheet still looks broadly plausible,
especially when grid lines are thin or the screenshot is downsampled. A
hard-coded verifier can instead read the workbook state directly and determine
exactly which cell contains which value.

\begin{figure}[H]
    \centering
    \includegraphics[width=0.98\linewidth]{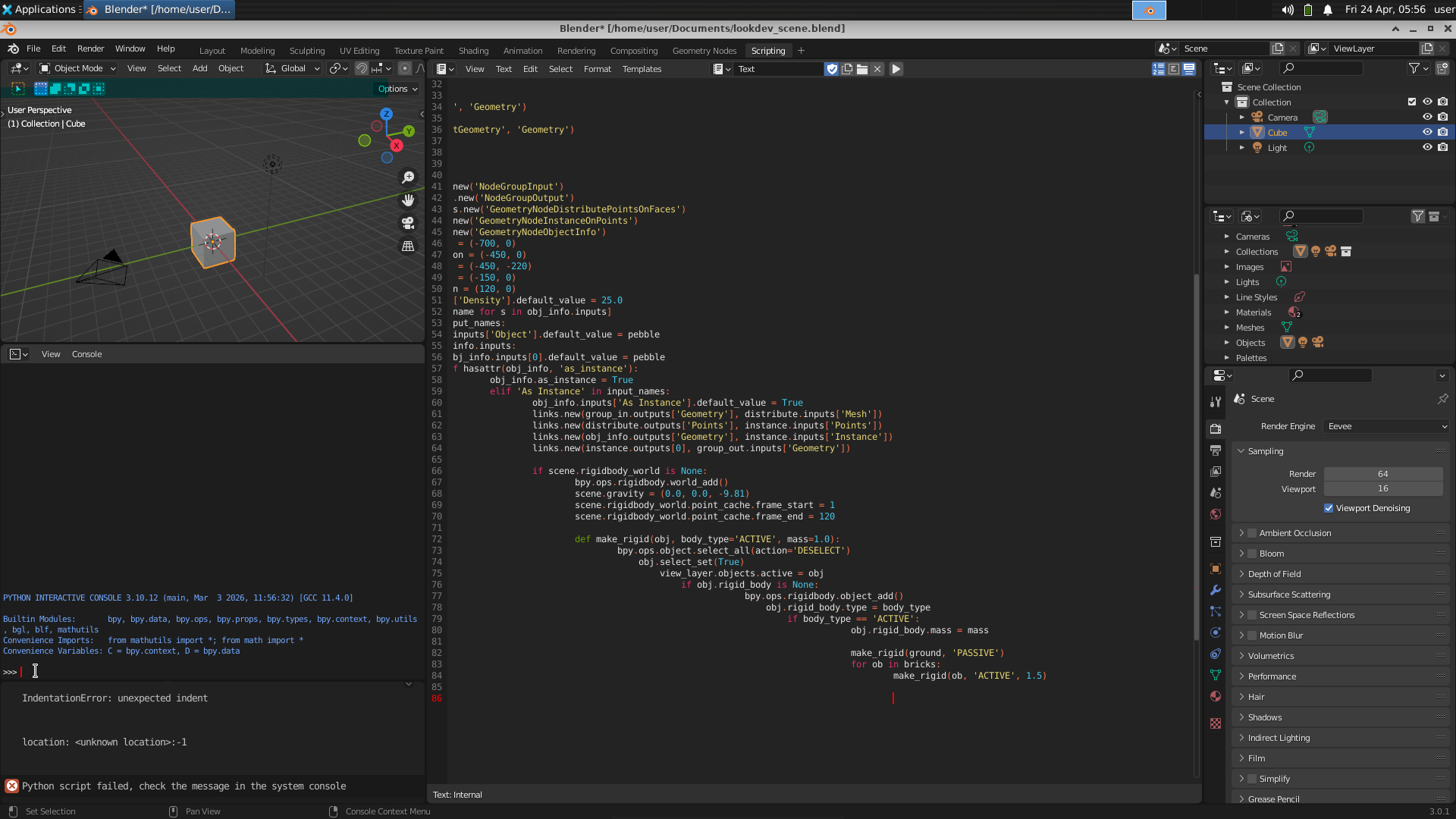}
    \caption{A terminal-heavy workflow where the decisive evidence lives in log
    lines, exit codes, and filesystem artifacts rather than in a clean final
    screenshot. Programmatic verifiers can inspect these sources directly,
    whereas screenshot-based judges only observe a narrow and potentially
    misleading window.}
    \label{fig:llm-judge-terminal}
\end{figure}

\paragraph{Failure mode 2: terminal-heavy tasks exceed screenshot context.}
For terminal-centric or mixed GUI-and-terminal workloads, the problem is not just
fine-grained visual ambiguity but limited observability. Figure~\ref{fig:llm-judge-terminal}
shows a case where the terminal contains the decisive evidence: an error line and
a missing output artifact. A screenshot captures only one scroll position and one
pane layout, so the judge must infer whether earlier logs, filesystem state, and
intermediate outputs are consistent with task completion.

\paragraph{Implication for OpenComputer.}
These examples motivate the separation of roles in OpenComputer. We use
LLM-as-judge as a flexible, high-level reference that helps detect verifier bugs,
underspecified criteria, and other pipeline issues during task construction. But
we reserve final scoring for hard-coded verifiers that inspect application-grounded
state directly. This choice makes rewards reproducible, auditable, and sensitive
to the exact success conditions that the benchmark is meant to evaluate.

%% file: Appendix/task_example.tex
\section{Examples of Generated Verifiable Tasks}
\label{app:generated-task-examples}

This section shows representative tasks generated by \ours across different
desktop applications. Each task is paired with executable verification criteria
that check the resulting application state, files, metadata, or persistent side
effects.

\begin{taskexample}{Zotero: Three-Level Collection Hierarchy}
\taskfield{Application} \texttt{zotero}

\taskfield{Task} Launch Zotero with the pre-seeded library at
\texttt{/home/user/Zotero/zotero.sqlite}. Under \texttt{My Library}, create a
top-level collection named \texttt{Papers}, a subcollection named
\texttt{Vision}, and a sub-subcollection named \texttt{Object Detection}.
Each collection must be a direct child of the previous level. Leave Zotero
running so the updated library state is saved.

\taskfield{Initial files}
\texttt{zotero.sqlite} at \texttt{/home/user/Zotero/zotero.sqlite}.

\taskfield{Representative verification checks}
\begin{compactitem}
    \item The collection \texttt{Papers} exists in the Zotero library.
    \item The collection \texttt{Vision} exists.
    \item The collection \texttt{Object Detection} exists.
    \item \texttt{Vision} is a direct child of \texttt{Papers}.
    \item \texttt{Object Detection} is a direct child of \texttt{Vision}.
\end{compactitem}
\end{taskexample}

\begin{taskexample}{MuseScore: Piano Solo to Piano Quartet}
\taskfield{Application} \texttt{musescore3}

\taskfield{Task} Open the score
\texttt{/home/user/Documents/piano\_sketch.mscz}, which contains a single
Piano part in C major, 4/4 time, and four measures of quarter notes. Arrange
the solo as a piano quartet by adding \texttt{Violin}, \texttt{Viola}, and
\texttt{Violoncello} parts below the Piano. Copy the Piano melody to the
Violin staff. On the Viola staff, enter one whole note per measure with pitches
\texttt{A3}, \texttt{D4}, \texttt{G3}, and \texttt{C4}. On the Violoncello
staff, enter a walking bass using four quarter notes per measure:
\texttt{C3}, \texttt{E3}, \texttt{G3}, \texttt{E3};
\texttt{F3}, \texttt{A3}, \texttt{C4}, \texttt{A3};
\texttt{G3}, \texttt{B3}, \texttt{D4}, \texttt{B3};
and \texttt{C3}, \texttt{E3}, \texttt{G3}, \texttt{C4}. Add dynamics
\texttt{mp}, \texttt{mf}, \texttt{p}, and \texttt{f} to measures 1--4,
respectively. Add staccato articulations to all four Violoncello notes in
measure 4 and nowhere else. Save the score, then export it as uncompressed
MusicXML to \texttt{/home/user/Documents/piano\_sketch.musicxml}.

\taskfield{Initial files}
\texttt{piano\_sketch.mscz} at
\texttt{/home/user/Documents/piano\_sketch.mscz}.

\taskfield{Representative verification checks}
\begin{compactitem}
    \item The original \texttt{.mscz} file still exists at the expected path.
    \item The saved score has exactly four parts: \texttt{Piano},
    \texttt{Violin}, \texttt{Viola}, and \texttt{Violoncello}.
    \item The score still has four measures and remains in 4/4 time.
    \item The score contains at least 52 notes, covering the original Piano
    melody, copied Violin melody, Viola whole notes, and Violoncello bass line.
    \item Measures 1--4 contain the expected dynamics:
    \texttt{mp}, \texttt{mf}, \texttt{p}, and \texttt{f}.
    \item Measure 4 contains staccato articulations.
    \item The total number of staccato articulations in the score is exactly 4.
    \item The exported MusicXML file exists at
    \texttt{/home/user/Documents/piano\_sketch.musicxml}.
    \item The exported MusicXML file contains exactly four parts.
\end{compactitem}
\end{taskexample}

\begin{taskexample}{LibreOffice Calc: Sales Commission Spreadsheet}
\taskfield{Application} \texttt{libreoffice\_calc}

\taskfield{Task} Open \texttt{/home/user/Documents/commissions.xlsx}. In the
\texttt{Sales} sheet, add bold headers \texttt{Commission Rate} and
\texttt{Commission Amount}. Use a nested \texttt{IF} formula to assign
commission rates based on monthly sales: 10\% for sales above 20,000, 8\% for
sales at least 10,000, and 5\% otherwise. Compute commission amounts for all
20 records, create a new \texttt{Commission Summary} sheet, and add formulas
for total sales, total commission, and average commission rate. Save the
workbook.

\taskfield{Initial files}
\texttt{commissions.xlsx} at \texttt{/home/user/Documents/commissions.xlsx}.

\taskfield{Representative verification checks}
\begin{compactitem}
    \item Cells \texttt{D1} and \texttt{E1} contain the expected bold headers.
    \item Cell \texttt{D2} contains an \texttt{IF} formula and evaluates to
    the expected commission rate.
    \item Cell \texttt{E2} references \texttt{C2} and \texttt{D2} and computes
    the expected commission amount.
    \item The copied formulas produce expected values for low-, mid-, and
    high-tier sales examples.
    \item The \texttt{Commission Summary} sheet exists.
    \item The summary sheet contains the expected labels and formulas for total
    sales, total commission, and average commission rate.
    \item The workbook is saved.
\end{compactitem}
\end{taskexample}

\begin{taskexample}{Obsidian: Recipe Vault Construction}
\taskfield{Application} \texttt{obsidian}

\taskfield{Task} Open the Obsidian vault at
\texttt{/home/user/Documents/RecipeVault}. Create folders for
\texttt{Italian}, \texttt{Asian}, and \texttt{Desserts}. Add four recipe notes
with YAML frontmatter, headings, ingredients, instructions, tags, and internal
links: \texttt{Carbonara.md}, \texttt{Cacio e Pepe.md},
\texttt{PadThai.md}, and \texttt{Tiramisu.md}. Finally, create a root
\texttt{Index.md} note linking to all recipes and tagged as an index.

\taskfield{Initial files}
None.

\taskfield{Representative verification checks}
\begin{compactitem}
    \item The \texttt{Italian}, \texttt{Asian}, and \texttt{Desserts} folders
    exist in the vault.
    \item Recipe notes contain the expected frontmatter fields, such as
    \texttt{cuisine}, \texttt{servings}, and \texttt{time}.
    \item \texttt{Carbonara} links to \texttt{Cacio e Pepe}, and
    \texttt{Cacio e Pepe} links back to \texttt{Carbonara}.
    \item \texttt{PadThai} contains the \texttt{\#noodles} tag.
    \item \texttt{Tiramisu} links to \texttt{Carbonara}.
    \item \texttt{Index.md} links to the expected recipes and contains the
    \texttt{\#index} tag.
    \item The vault contains exactly five notes.
\end{compactitem}
\end{taskexample}

\begin{taskexample}{Blender: Parent-Child Rig Hierarchy}
\taskfield{Application} \texttt{blender}

\taskfield{Task} Open the Blender file
\texttt{/home/user/Documents/rig\_hierarchy.blend}, which starts empty. Add an
Empty and rename it to \texttt{Rig}. Add a cube named \texttt{Torso}, a UV
sphere named \texttt{Head}, and a cylinder named \texttt{Arm}. Parent
\texttt{Torso} and \texttt{Arm} to \texttt{Rig}, and parent \texttt{Head} to
\texttt{Torso}, making \texttt{Head} a grandchild of \texttt{Rig}. Save the
file.

\taskfield{Initial files}
\texttt{rig\_hierarchy.blend} at
\texttt{/home/user/Documents/rig\_hierarchy.blend}.

\taskfield{Representative verification checks}
\begin{compactitem}
    \item The saved Blender file exists at the expected path.
    \item The object \texttt{Rig} exists and has type \texttt{EMPTY}.
    \item The objects \texttt{Torso}, \texttt{Head}, and \texttt{Arm} exist and
    have mesh object types.
    \item \texttt{Torso} is parented to \texttt{Rig}.
    \item \texttt{Arm} is parented to \texttt{Rig}.
    \item \texttt{Head} is parented to \texttt{Torso}.
\end{compactitem}
\end{taskexample}

%% file: references.bib
@article{fang2025towards,
  title={Towards general agentic intelligence via environment scaling},
  author={Fang, Runnan and Cai, Shihao and Li, Baixuan and Wu, Jialong and Li, Guangyu and Yin, Wenbiao and Wang, Xinyu and Wang, Xiaobin and Su, Liangcai and Zhang, Zhen and others},
  journal={arXiv preprint arXiv:2509.13311},
  year={2025}
}

@article{wang2026agent,
  title={Agent world model: Infinity synthetic environments for agentic reinforcement learning},
  author={Wang, Zhaoyang and Xu, Canwen and Liu, Boyi and Wang, Yite and Han, Siwei and Yao, Zhewei and Yao, Huaxiu and He, Yuxiong},
  journal={arXiv preprint arXiv:2602.10090},
  year={2026}
}

@article{li2025simulating,
  title={Simulating environments with reasoning models for agent training},
  author={Li, Yuetai and Inan, Huseyin A and Yue, Xiang and Chen, Wei-Ning and Wutschitz, Lukas and Kulkarni, Janardhan and Poovendran, Radha and Sim, Robert and Rajmohan, Saravan},
  journal={arXiv preprint arXiv:2511.01824},
  year={2025}
}

@article{zhang2026infiniteweb,
  title={InfiniteWeb: Scalable Web Environment Synthesis for GUI Agent Training},
  author={Zhang, Ziyun and Wang, Zezhou and Zhang, Xiaoyi and Guo, Zongyu and Li, Jiahao and Li, Bin and Lu, Yan},
  journal={arXiv preprint arXiv:2601.04126},
  year={2026}
}

@article{cao2026gui,
  title={GUI-GENESIS: Automated Synthesis of Efficient Environments with Verifiable Rewards for GUI Agent Post-Training},
  author={Cao, Yuan and Ran, Dezhi and Wu, Mengzhou and Guo, Yuzhe and Chen, Xin and Li, Ang and Cao, Gang and Zhi, Gong and Yu, Hao and Li, Linyi and others},
  journal={arXiv preprint arXiv:2602.14093},
  year={2026}
}

@article{aggarwal2026gym,
  title={Gym-Anything: Turn any Software into an Agent Environment},
  author={Aggarwal, Pranjal and Neubig, Graham and Welleck, Sean},
  journal={arXiv preprint arXiv:2604.06126},
  year={2026}
}

@article{zhu2026termigen,
  title={TermiGen: High-Fidelity Environment and Robust Trajectory Synthesis for Terminal Agents},
  author={Zhu, Kaijie and Nie, Yuzhou and Li, Yijiang and Huang, Yiming and Wu, Jialian and Liu, Jiang and Sun, Ximeng and Yin, Zhenfei and Wang, Lun and Liu, Zicheng and others},
  journal={arXiv preprint arXiv:2602.07274},
  year={2026}
}

@article{zhao2026immersion,
  title={Immersion in the GitHub Universe: Scaling Coding Agents to Mastery},
  author={Zhao, Jiale and Chen, Guoxin and Meng, Fanzhe and Li, Minghao and Chen, Jie and Xu, Hui and Sun, Yongshuai and Zhao, Wayne Xin and Song, Ruihua and Zhang, Yuan and others},
  journal={arXiv preprint arXiv:2602.09892},
  year={2026}
}

@misc{openai2026gpt54,
  title        = {GPT-5.4 Model},
  author       = {{OpenAI}},
  year         = {2026},
  howpublished = {\url{https://developers.openai.com/api/docs/models/gpt-5.4}}
}

@misc{anthropic2026sonnet46,
  title        = {Introducing Claude Sonnet 4.6},
  author       = {{Anthropic}},
  year         = {2026},
  howpublished = {\url{https://www.anthropic.com/news/claude-sonnet-4-6}}
}

@misc{moonshot2026kimi26,
  title        = {Kimi K2.6},
  author       = {{Moonshot AI}},
  year         = {2026},
  howpublished = {\url{https://huggingface.co/moonshotai/Kimi-K2.6}}
}

@misc{google2025gemini3flash,
  title        = {Gemini 3 Flash: Frontier Intelligence Built for Speed},
  author       = {{Google}},
  year         = {2025},
  howpublished = {\url{https://blog.google/products-and-platforms/products/gemini/gemini-3-flash/}}
}

@misc{qwen2026qwen35,
  title        = {Qwen3.5 Model Family},
  author       = {{Qwen Team}},
  year         = {2026},
  howpublished = {\url{https://huggingface.co/collections/Qwen/qwen35}}
}

@article{xue2026evocua,
  title={Evocua: Evolving computer use agents via learning from scalable synthetic experience},
  author={Xue, Taofeng and Peng, Chong and Huang, Mianqiu and Guo, Linsen and Han, Tiancheng and Wang, Haozhe and Wang, Jianing and Zhang, Xiaocheng and Yang, Xin and Zhao, Dengchang and others},
  journal={arXiv preprint arXiv:2601.15876},
  year={2026}
}

@article{xu2026mobile,
  title={Mobile-agent-v3. 5: Multi-platform fundamental gui agents},
  author={Xu, Haiyang and Zhang, Xi and Liu, Haowei and Wang, Junyang and Zhu, Zhaozai and Zhou, Shengjie and Hu, Xuhao and Gao, Feiyu and Cao, Junjie and Wang, Zihua and others},
  journal={arXiv preprint arXiv:2602.16855},
  year={2026}
}

@article{xie2024osworld,
  title={Osworld: Benchmarking multimodal agents for open-ended tasks in real computer environments},
  author={Xie, Tianbao and Zhang, Danyang and Chen, Jixuan and Li, Xiaochuan and Zhao, Siheng and Cao, Ruisheng and Hua, Toh J and Cheng, Zhoujun and Shin, Dongchan and Lei, Fangyu and others},
  journal={Advances in Neural Information Processing Systems},
  volume={37},
  pages={52040--52094},
  year={2024}
}

@article{deng2023mind2web,
  title={Mind2web: Towards a generalist agent for the web},
  author={Deng, Xiang and Gu, Yu and Zheng, Boyuan and Chen, Shijie and Stevens, Sam and Wang, Boshi and Sun, Huan and Su, Yu},
  journal={Advances in Neural Information Processing Systems},
  volume={36},
  pages={28091--28114},
  year={2023}
}

@article{rawles2023androidinthewild,
  title={Androidinthewild: A large-scale dataset for android device control},
  author={Rawles, Christopher and Li, Alice and Rodriguez, Daniel and Riva, Oriana and Lillicrap, Timothy},
  journal={Advances in Neural Information Processing Systems},
  volume={36},
  pages={59708--59728},
  year={2023}
}

@article{bonatti2024windows,
  title={Windows agent arena: Evaluating multi-modal os agents at scale},
  author={Bonatti, Rogerio and Zhao, Dan and Bonacci, Francesco and Dupont, Dillon and Abdali, Sara and Li, Yinheng and Lu, Yadong and Wagle, Justin and Koishida, Kazuhito and Bucker, Arthur and others},
  journal={arXiv preprint arXiv:2409.08264},
  year={2024}
}

@article{zhou2023webarena,
  title={Webarena: A realistic web environment for building autonomous agents},
  author={Zhou, Shuyan and Xu, Frank F and Zhu, Hao and Zhou, Xuhui and Lo, Robert and Sridhar, Abishek and Cheng, Xianyi and Ou, Tianyue and Bisk, Yonatan and Fried, Daniel and others},
  journal={arXiv preprint arXiv:2307.13854},
  year={2023}
}

@inproceedings{koh2024visualwebarena,
  title={Visualwebarena: Evaluating multimodal agents on realistic visual web tasks},
  author={Koh, Jing Yu and Lo, Robert and Jang, Lawrence and Duvvur, Vikram and Lim, Ming and Huang, Po-Yu and Neubig, Graham and Zhou, Shuyan and Salakhutdinov, Russ and Fried, Daniel},
  booktitle={Proceedings of the 62nd Annual Meeting of the Association for Computational Linguistics (Volume 1: Long Papers)},
  pages={881--905},
  year={2024}
}

@article{drouin2024workarena,
  title={Workarena: How capable are web agents at solving common knowledge work tasks?},
  author={Drouin, Alexandre and Gasse, Maxime and Caccia, Massimo and Laradji, Issam H and Del Verme, Manuel and Marty, Tom and Boisvert, L{\'e}o and Thakkar, Megh and Cappart, Quentin and Vazquez, David and others},
  journal={arXiv preprint arXiv:2403.07718},
  year={2024}
}

@article{rawles2024androidworld,
  title={Androidworld: A dynamic benchmarking environment for autonomous agents},
  author={Rawles, Christopher and Clinckemaillie, Sarah and Chang, Yifan and Waltz, Jonathan and Lau, Gabrielle and Fair, Marybeth and Li, Alice and Bishop, William and Li, Wei and Campbell-Ajala, Folawiyo and others},
  journal={arXiv preprint arXiv:2405.14573},
  year={2024}
}

@inproceedings{agasheagent,
  title={Agent S: An Open Agentic Framework that Uses Computers Like a Human},
  author={Agashe, Saaket and Han, Jiuzhou and Gan, Shuyu and Yang, Jiachen and Li, Ang and Wang, Xin Eric},
  booktitle={The Thirteenth International Conference on Learning Representations}
}

@inproceedings{nguyen2025gui,
  title={Gui agents: A survey},
  author={Nguyen, Dang and Chen, Jian and Wang, Yu and Wu, Gang and Park, Namyong and Hu, Zhengmian and Lyu, Hanjia and Wu, Junda and Aponte, Ryan and Xia, Yu and others},
  booktitle={Findings of the Association for Computational Linguistics: ACL 2025},
  pages={22522--22538},
  year={2025}
}

@article{agashe2025agent,
  title={Agent s2: A compositional generalist-specialist framework for computer use agents},
  author={Agashe, Saaket and Wong, Kyle and Tu, Vincent and Yang, Jiachen and Li, Ang and Wang, Xin Eric},
  journal={arXiv preprint arXiv:2504.00906},
  year={2025}
}

@article{song2025coact,
  title={Coact-1: Computer-using agents with coding as actions},
  author={Song, Linxin and Dai, Yutong and Prabhu, Viraj and Zhang, Jieyu and Shi, Taiwei and Li, Li and Li, Junnan and Savarese, Silvio and Chen, Zeyuan and Zhao, Jieyu and others},
  journal={arXiv preprint arXiv:2508.03923},
  year={2025}
}

@inproceedings{liu2023g,
  title={G-eval: NLG evaluation using gpt-4 with better human alignment},
  author={Liu, Yang and Iter, Dan and Xu, Yichong and Wang, Shuohang and Xu, Ruochen and Zhu, Chenguang},
  booktitle={Proceedings of the 2023 conference on empirical methods in natural language processing},
  pages={2511--2522},
  year={2023}
}

@article{zheng2023judging,
  title={Judging llm-as-a-judge with mt-bench and chatbot arena},
  author={Zheng, Lianmin and Chiang, Wei-Lin and Sheng, Ying and Zhuang, Siyuan and Wu, Zhanghao and Zhuang, Yonghao and Lin, Zi and Li, Zhuohan and Li, Dacheng and Xing, Eric and others},
  journal={Advances in neural information processing systems},
  volume={36},
  pages={46595--46623},
  year={2023}
}

@inproceedings{wang2024large,
  title={Large language models are not fair evaluators},
  author={Wang, Peiyi and Li, Lei and Chen, Liang and Cai, Zefan and Zhu, Dawei and Lin, Binghuai and Cao, Yunbo and Kong, Lingpeng and Liu, Qi and Liu, Tianyu and others},
  booktitle={Proceedings of the 62nd Annual Meeting of the Association for Computational Linguistics (Volume 1: Long Papers)},
  pages={9440--9450},
  year={2024}
}

@article{xu2024agenttrek,
  title={Agenttrek: Agent trajectory synthesis via guiding replay with web tutorials},
  author={Xu, Yiheng and Lu, Dunjie and Shen, Zhennan and Wang, Junli and Wang, Zekun and Mao, Yuchen and Xiong, Caiming and Yu, Tao},
  journal={arXiv preprint arXiv:2412.09605},
  year={2024}
}

@article{he2024pc,
  title={PC Agent: While You Sleep, AI Works--A Cognitive Journey into Digital World},
  author={He, Yanheng and Jin, Jiahe and Xia, Shijie and Su, Jiadi and Fan, Runze and Zou, Haoyang and Hu, Xiangkun and Liu, Pengfei},
  journal={arXiv preprint arXiv:2412.17589},
  year={2024}
}

@article{dai2025scuba,
  title={SCUBA: Salesforce Computer Use Benchmark},
  author={Dai, Yutong and Ramakrishnan, Krithika and Gu, Jing and Fernandez, Matthew and Luo, Yanqi and Prabhu, Viraj and Hu, Zhenyu and Savarese, Silvio and Xiong, Caiming and Chen, Zeyuan and others},
  journal={arXiv preprint arXiv:2509.26506},
  year={2025}
}

@misc{hkuds2026clianything,
  title        = {{CLI-Anything}: Making ALL Software Agent-Native},
  author       = {{HKUDS}},
  year         = {2026},
  howpublished = {\url{https://github.com/HKUDS/CLI-Anything}},
  note         = {GitHub repository. Accessed: 2026-05-02}
}

@inproceedings{kim2024prometheus,
  title={Prometheus: Inducing fine-grained evaluation capability in language models},
  author={Kim, Seungone and Shin, Jay and Jang, Joel and Longpre, Shayne and Lee, Hwaran and Yun, Sangdoo and Shin, Ryan and Kim, Sungdong and Thorne, James and Seo, Minjoon and others},
  booktitle={International Conference on Learning Representations},
  volume={2024},
  pages={29927--29962},
  year={2024}
}

@article{song2025bearcubs,
  title={Bearcubs: A benchmark for computer-using web agents},
  author={Song, Yixiao and Thai, Katherine and Pham, Chau Minh and Chang, Yapei and Nadaf, Mazin and Iyyer, Mohit},
  journal={arXiv preprint arXiv:2503.07919},
  year={2025}
}

@inproceedings{ye2026realwebassist,
  title={Realwebassist: A benchmark for long-horizon web assistance with real-world users},
  author={Ye, Suyu and Shi, Haojun and Shih, Darren and Yun, Hyokun and Roosta, Tanya G and Shu, Tianmin},
  booktitle={Proceedings of the AAAI Conference on Artificial Intelligence},
  volume={40},
  number={40},
  pages={34441--34449},
  year={2026}
}

@inproceedings{thakur2025judging,
  title={Judging the judges: Evaluating alignment and vulnerabilities in llms-as-judges},
  author={Thakur, Aman Singh and Choudhary, Kartik and Ramayapally, Venkat Srinik and Vaidyanathan, Sankaran and Hupkes, Dieuwke},
  booktitle={Proceedings of the Fourth Workshop on Generation, Evaluation and Metrics (GEM$^2$)},
  pages={404--430},
  year={2025}
}

@inproceedings{li2025generation,
  title={From generation to judgment: Opportunities and challenges of llm-as-a-judge},
  author={Li, Dawei and Jiang, Bohan and Huang, Liangjie and Beigi, Alimohammad and Zhao, Chengshuai and Tan, Zhen and Bhattacharjee, Amrita and Jiang, Yuxuan and Chen, Canyu and Wu, Tianhao and others},
  booktitle={Proceedings of the 2025 Conference on Empirical Methods in Natural Language Processing},
  pages={2757--2791},
  year={2025}
}

@article{sumyk2026cuaaudit,
  title={CUAAudit: Meta-Evaluation of Vision-Language Models as Auditors of Autonomous Computer-Use Agents},
  author={Sumyk, Marta and Kosovan, Oleksandr},
  journal={arXiv preprint arXiv:2603.10577},
  year={2026}
}

@article{cui2026agentic,
  title={Agentic Reward Modeling: Verifying GUI Agent via Online Proactive Interaction},
  author={Cui, Chaoqun and Huang, Jing and Wang, Shijing and Zheng, Liming and Kong, Qingchao and Zeng, Zhixiong},
  journal={arXiv preprint arXiv:2602.00575},
  year={2026}
}
